%% file: coling_latex.tex
\pdfoutput=1

\documentclass[11pt]{article}

\usepackage[preprint]{coling}

\usepackage{times}
\usepackage{latexsym}

\usepackage[T1]{fontenc}

\usepackage[utf8]{inputenc}

\usepackage{microtype}

\usepackage{inconsolata}

\usepackage{graphicx}
\usepackage{amsmath}
\usepackage{booktabs}
\usepackage{multirow}
\usepackage{makecell}
\usepackage{hyperref}
\usepackage{svg}
\usepackage{xcolor}

\DeclareRobustCommand{\legendsquare}[1]{%
  \textcolor{#1}{\rule{1ex}{1ex}}%
}

\newcommand{\creamemoji}{
    \includegraphics[width=12pt]{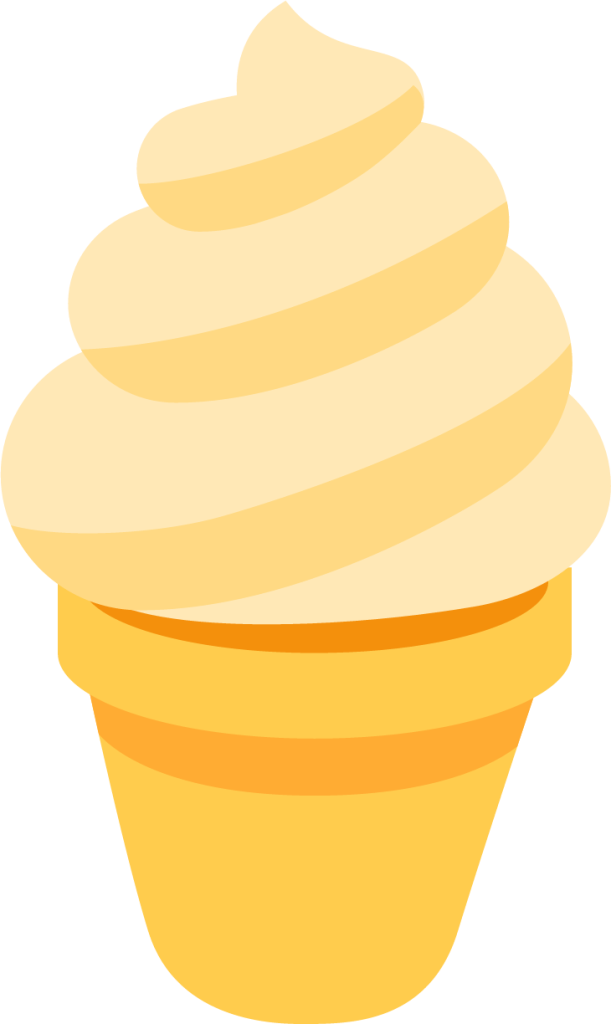}
}

\title{CREAM \creamemoji: Comparison-Based Reference-Free ELO-Ranked Automatic Evaluation for Meeting Summarization}

\author{
    \textbf{Ziwei Gong\textsuperscript{1,2}},
    \textbf{Lin Ai\textsuperscript{1,2}},
    \textbf{Harshsaiprasad Deshpande\textsuperscript{1}},
    \textbf{Alexander Johnson\textsuperscript{1}}, \\
    \textbf{Emmy Phung\textsuperscript{1}},
    \textbf{Zehui Wu \textsuperscript{2}},
    \textbf{Ahmad Emami\textsuperscript{1}},
    \textbf{Julia Hirschberg\textsuperscript{2}}
    \\
    \\
    \textsuperscript{1}Machine Learning Center of Excellence, JPMorgan Chase \& Co.\\
    \textsuperscript{2}Department of Computer Science, Columbia University
    \\
    \{lin.ai, sara.ziweigong, julia\}@cs.columbia.edu, 
}

\begin{document}
\maketitle
\begin{abstract}

Large Language Models (LLMs) have spurred interest in automatic evaluation methods for summarization, offering a faster, more cost-effective alternative to human evaluation. However, existing methods often fall short when applied to complex tasks like long-context summarizations and dialogue-based meeting summarizations. In this paper, we introduce CREAM (\underline{C}omparison-based \underline{R}eference-free \underline{E}lo-ranked \underline{A}utomatic evaluation for \underline{M}eeting summarization), a novel framework that addresses the unique challenges of evaluating meeting summaries. CREAM leverages a combination of chain-of-thought reasoning and key facts alignment to assess conciseness and completeness of model-generated summaries without requiring reference. By employing an ELO ranking system, our approach provides a robust mechanism for comparing the quality of different models or prompt configurations.

\end{abstract}

\section{Introduction}

The rapid advancement of Large Language Models (LLMs) has significantly influenced the field of automatic evaluation for text summarization. LLMs offer the potential to streamline the evaluation process, making it faster and more cost-effective compared to traditional human evaluation \cite{liu-etal-2023-g, wang-etal-2023-chatgpt}. However, despite the progress in automatic evaluation techniques, existing methods primarily target general-purpose summarization tasks, which typically involve shorter, more straightforward text inputs, which may not be adaptable to long-context settings \cite{chang2024booookscore}. Meeting and dialogue summarization is crucial in NLP applications, especially in industrial settings, where accurate summaries of lengthy meetings enhance decision-making, communication, and knowledge retention. They convert discussions into actionable insights, supporting project management, compliance, and strategic planning, ultimately boosting productivity and efficiency. 


Long-context documents and dialogue-based meeting summarizations present unique challenges that are not adequately addressed by current automatic evaluation metrics \cite{chang2024booookscore}. These tasks require a deeper understanding of the content, context, and flow of information, which often leads to limitations in the applicability of general-purpose metrics. For example, a well-documented issue in current automatic evaluation methods, known as the "middle curse" \cite{liu-etal-2024-lost, ravaut-etal-2024-context},  exacerbates this problem. This phenomenon refers to the tendency of models to perform well at the beginning and end of a summary while neglecting or misrepresenting information in the middle sections. This raises questions about the effectiveness of existing LLM-based evaluators for meeting summarization, a hypothesis we thoroughly examine in this work. Our investigation reveals that current methods do not effectively evaluate meeting summarizations, prompting the need for a more specialized approach.

\begin{figure*}[ht]
\centering
\includegraphics[width=0.93\textwidth]{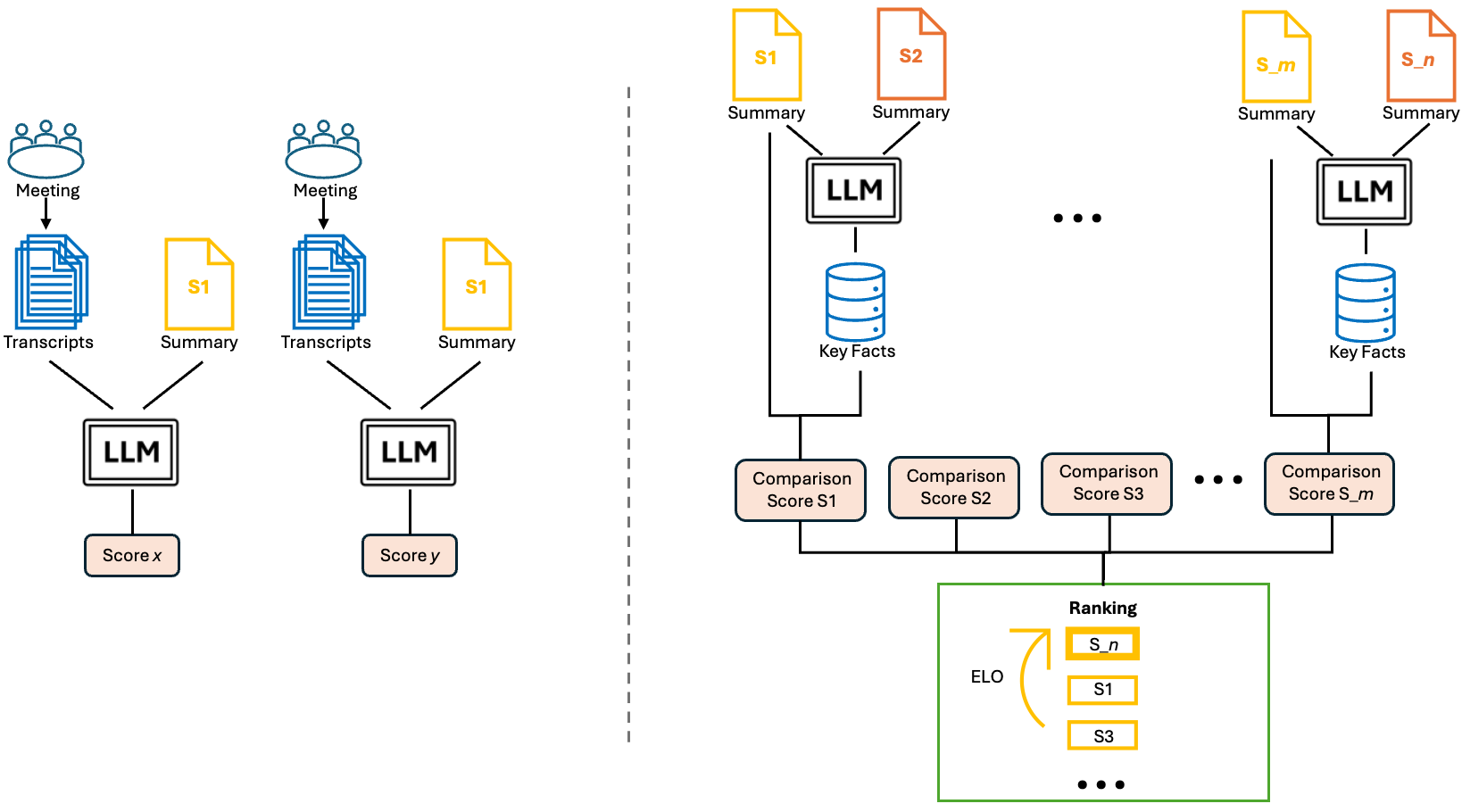}
\caption{Illustrations of current evaluation framework (left) and the CREAM framework (right). On the left, traditional methods independently summarize and score each meeting transcript against reference texts. On the right, CREAM distills candidate summary pairs into key facts, conducts pair wise comparison of summary pairs, and then uses an Elo rating system to rank summaries based on their relative quality.}
\label{tab:CREAM}
\end{figure*}

In this paper, we address this gap by developing a new evaluation framework tailored specifically for meeting summarization.We propose CREAM (\underline{C}omparison-based \underline{R}eference-free \underline{E}lo-ranked \underline{A}utomatic evaluation for \underline{M}eeting summarization), a novel system designed to fill the gaps in specialized and customizable evaluation for meeting summaries as illustrated in Figure \ref{tab:CREAM}. Our research addresses the following key questions:
\begin{enumerate}
\item Do current LLM-based automatic evaluators work effectively for meeting summarization? (Our research shows that they do not.)
\vspace{-0.2cm}
\item How can we design an efficient, reference-free, automatic evaluator for meeting summarization? (We propose comparison-based CREAM framework.)
\vspace{-0.2cm}
\item How to compare different models using comparison-based metrics? (We introduce an ELO ranking system.)
\end{enumerate}

Our results highlight the limitations of current LLM-based evaluators and demonstrate the effectiveness of our specialized framework, CREAM, with its novel comparison-based Elo ranking method for summarization evaluation. We benchmark various GPT models against our framework and find that GPT-4o excels in completeness, GPT-4 in conciseness, but all struggle to find a balance between completeness and conciseness.

\section{Related Work}
\subsection{Reference-based Evaluation Metrics}

Meeting summarization commonly uses reference-based metrics to assess summary quality.
Ngram-based metrics like ROUGE \cite{lin-2004-rouge} measure n-gram overlap between generated and reference summaries, with variants such as ROUGE-1 (unigram), ROUGE-2 (bigram), ROUGE-L (longest common subsequence). BLEU \cite{10.3115/1073083.1073135}, originally designed to evaluate machine translation, calculates n-gram precision with a brevity penalty and is adaptable for summarization.
Embedding-based metrics include BERTScore \cite{bert-score}, which uses BERT embeddings to assess semantic overlap at the token level, providing more nuanced comparisons. BARTScore \cite{NEURIPS2021_e4d2b6e6} evaluates the generation likelihood of reference text from the source or vice versa using BART.

While widely used, these similarity-based metrics often fail to capture multidimensional aspects of text quality, such as factuality, conciseness, and completeness, which are critical in human evaluations. This highlights the need for more advanced evaluation frameworks for meeting summarization.

\subsection{LLM-based Evaluation and Metrics}
LLM-based evaluation methods, like G-Eval \cite{liu-etal-2023-g} and FineSurE \cite{song2024finesure}, are improving alignment with human judgments in text summarization. G-Eval uses GPT-4 with a Chain-of-Thought approach, achieving strong correlations with human evaluations, though it may bias toward LLM-generated texts. FineSurE offers fine-grained assessments, enhancing evaluations of faithfulness, completeness, and conciseness.

However, the application of these LLM-based methods to meeting summarization, which involves long-context dialogue, remains under-explored. It's unclear if these generalized methods are fully applicable, indicating a need for further adaptation.

\subsection{Elo Rating in NLG Model Evaluation}

The Elo rating system, originally introduced by Arpad Elo in 1967 \cite{elo}, is widely used for ranking competitors in games like chess and has recently been adapted for evaluating natural language generation (NLG) systems. 

\citet{rackauckas2024evaluatingragfusionrageloautomated} applied an Elo-based framework, RAGElo, to evaluate Retrieval-Augmented Generation (RAG) systems, finding that it aligns well with expert annotations and effectively handles domain-specific challenges. Similarly, \citet{boubdir-etal-2023-elo} examined the robustness of Elo ratings for LLMs, emphasizing the need for careful calibration to ensure reliable rankings, particularly in non-transitive scenarios.

In a different approach, \citet{zhao2024autoarenallmsautomating} introduced Auto-Arena, an automated Elo-based evaluation system where LLMs engage in peer battles, which showed high correlation with human preferences, suggesting it as a viable alternative to human evaluation. Additionally, \citet{zhang2023llmevalpreliminarystudyevaluate} explored various ranking systems, including Elo, and analyzed their effectiveness for LLM evaluation.

Our work extends these applications by using Elo to rank models based on comparison-based evaluations of summarization conciseness and completeness, offering a nuanced and reliable framework specifically for meeting summarizations.

\begin{figure*}[ht]
\centering
\includegraphics[width=\textwidth]{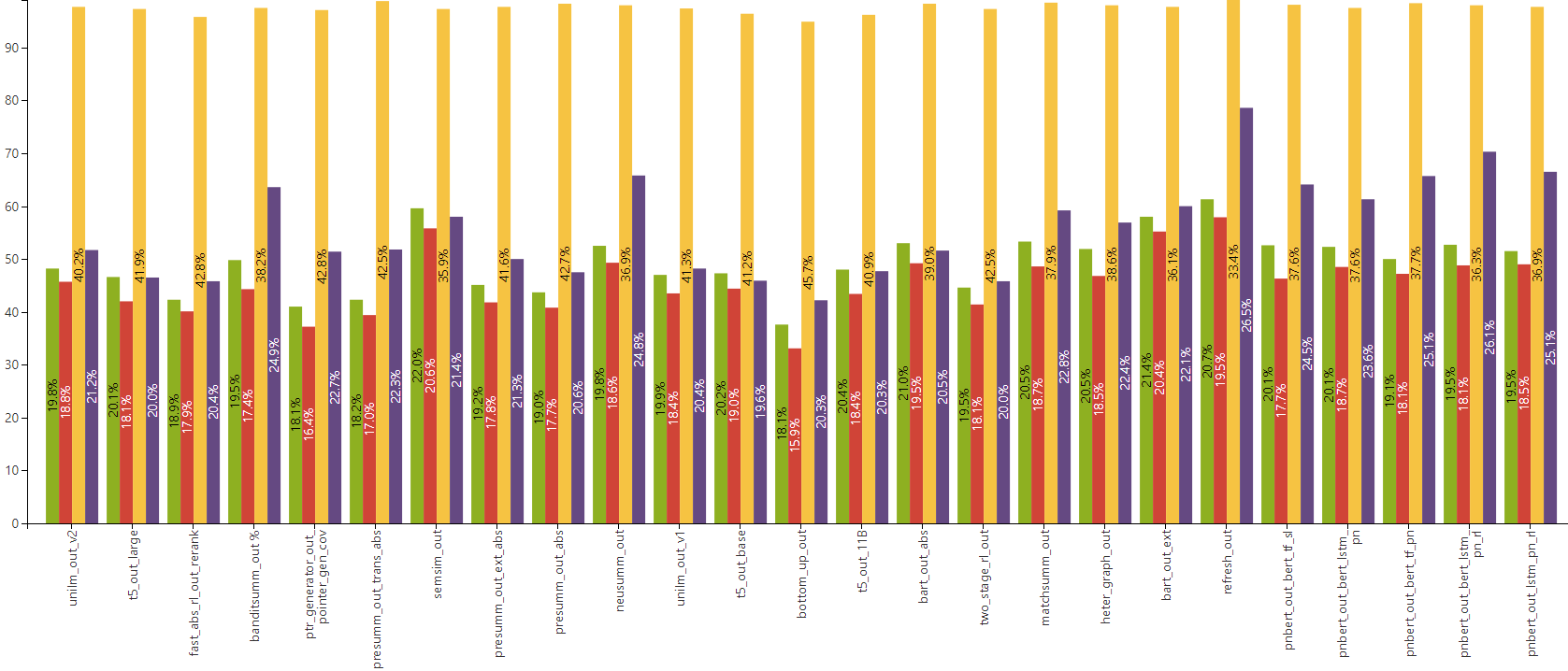}
\caption{Baseline completeness scores on the REALSum dataset. The $x$-axis shows different summarization models, and $y$-axis shows the completeness score under various settings: \legendsquare{green} represents scores using human annotations for both key facts and alignment. \legendsquare{red} shows scores with human-annotated key facts and machine alignment. \legendsquare{yellow} indicates scores with machine-annotated key facts from the summary and machine alignment. \legendsquare{violet} shows scores with machine-annotated key facts from the transcript and machine alignment.}
\label{tab:REALSum_completeness}
\end{figure*}

\input{table/completeness-conciseness-baseline}

\section{Challenges and Limitations of Current Approaches}
In order to answer the question, \textit{"Do current evaluators work effectively for meeting summarization?"}, we experiment with the state-of-the-art frameworks \cite{liu-etal-2023-g, song2024finesure} for automatic summary evaluation and identify several issues. 

The existing LLM-based methods, as illustrated on the left side of Figure \ref{tab:CREAM}, take individual meeting transcripts as input and the a separate summary for each transcript, and output an absolute score of the summary against reference texts under different metrics. This process involves comparing each generated summary to reference documents to assess their quality through CoT process, and takes human written summary as gold standards. This method primarily relies on comparing summaries to reference texts, often has a bias towards self-generated content. A detailed description of these methods and prompts is provided in Appendix \ref{sec:appendix_metrics}. 


Our experiments with existing methods show that: 1) current LLM-based evaluators often provide inaccurate completeness and conciseness scores for long-context documents, showing a high self-bias, due to challenges in extracting key facts from complex scenarios; 2) models like GPT-4o frequently produce false positives in real-world complex dialogue, indicating in industrial applications the focus should prioritize completeness and conciseness in evaluation frameworks.

\subsection{Experiment Setup}
\label{sec:setup}
\paragraph{Datasets}
We select both public and private datasets which collectively cover a range of summarization-related tasks: factuality assessment, key-fact extraction, and summarization.

FRANK \cite{pagnoni-etal-2021-understanding}, containing 2,246 summaries with sentence-level faithfulness annotation, and REALSumm \cite{bhandari-etal-2020-evaluating}, containing 2,500 summaries with key facts level alignment, are both factuality-focused datasets based on CNNDM \cite{hermann2015teaching} and XSUM datasets \cite{narayan-etal-2018-dont}. We select both datasets as we need both types of annotations for key-fact extraction and factuality evaluation.

QMSum \cite{zhong-etal-2021-qmsum} is a query-based, long meeting summarization dataset derived from the AMI \cite{10.1007/11677482_3}, ICSI \cite{1198793}, and Parliamentary committee meeting datasets. It includes 1,808 query-summary pairs from 232 meetings with average number of 556.8 turns across diverse domains including product, academic, and committee meetings. 

Our Internal Zoom Meeting Summarization (IZMS) data consists of 139 examples from internal Zoom meetings, used to assess factuality and key-fact extraction. This dataset includes transcripts from 10 product meetings and an additional 40 meetings, providing real-world data for evaluation.

\paragraph{LLMs used in Evaluation} 
We conduct experiments using GPT-based models from OpenAI \cite{openai2024gpt4technicalreport}, including GPT-4-omni (gpt-4o-2024-05-13), GPT-4-turbo (gpt-4-turbo-2024-04-09), and GPT-3.5 (gpt-35-turbo-16k-0613). Each model is evaluated for performance in generating and summarizing text, allowing us to assess the strengths and limitations of different versions in summarization tasks. We clear the history for each evaluation instance, following precedents \cite{song2024finesure, shen-etal-2023-large}.

\paragraph{Metrics} 

We evaluate LLM reliability on three key metrics: completeness, conciseness, and faithfulness, using definitions and scoring methods consistent with recent research \cite{song2024finesure}. Detailed descriptions and formulas are provided in Appendix \ref{sec:appendix_metrics}.

\textit{Completeness} assesses the extent to which the summarizer includes all key facts\footnote{A key fact refers to a concise sentence conveying a single key information, comprising at most 2-3 entities \cite{bhandari-etal-2020-evaluating, song2024finesure}.} from the input text in the output summary. 

\textit{Conciseness} measures the summarizer's ability to avoid incorporating information outside the key facts in the output, maintaining focus and brevity.

\textit{Faithfulness} evaluates whether the summarizer accurately represents the information in the input text without manipulating it (intrinsic) or adding information that is not directly inferable from the input text (extrinsic).

\subsection{Experiments}
\textbf{For both completeness and conciseness}, LLM evaluators show low correlation with human preference and high self-bias in long-context dialogues like meeting summaries, although perform well in evaluating shorter summaries.

\paragraph{Completeness}
LLMs evaluators like GPT-4o can accurately extract key facts and compare summaries with high correlation to human evaluations in shorter summarization tasks like those in REALSum, which involve news-style articles with summaries averaging $3-5$ sentences. We evaluate LLMs’ ability to measure completeness when given different type of reference by extract key facts and align them with summaries generated by various baseline models, comparing their performance against human-annotated scores. As shown in Figure \ref{tab:REALSum_completeness}, 
GPT-4o showed a high correlation with human scores when using human-annotated key facts as references (rank correlation of $0.95$, Pearson's r), and also similarly positive correlation when extracting key facts from entire articles for comparison with summaries.

However, these evaluators struggle with accurately evaluating completeness in long-context dialogues, such as meeting summaries from the QMSum dataset with average of $556.8$ turns. As shown in Table \ref{tab:completeness-conciseness-baseline-table}, our evaluation revealed that all versions of GPT have a weak correlation (Pearson's r $= 0.5$, ) with gold standard human scores with a strong self-bias. Both GPT-4 and GPT-3.5 tend to give near-perfect scores for summaries generated by LLMs, with GPT-4o performing only slightly better but still showing a bias towards its own generated content. 

\paragraph{Conciseness}
Similar to completeness, these evaluators have a weak correlation with gold standard human scores with a strong self-bias when evaluating conciseness. As shown in Table \ref{tab:completeness-conciseness-baseline-table}, LLMs are far less effective for long-context dialogues, and performance without gold reference tends to result in an overestimation of conciseness.


\paragraph{Factuality}
\label{sec:Factualityfindings}

Current evaluators struggle with accurately identifying factual errors in meeting summaries. We assessed whether current LLM-based evaluators (GPT-4o) are effective in evaluating factuality. We reproduce results on the FRANK dataset, which has fine-grained annotations
of sentence-level factuality error types. We achieve a balanced accuracy of $61.5\%$ at the sentence level. Summary-level correlations with human judgments were moderate, with Pearson ($0.38$) and Spearman ($0.35$) correlations. At system level, the rank correlation is perfect (Spearman'r of $1.0$). 

However, when applying the same method to our IZMS data, the results are less promising. The summary-level correlations were low (Pearson = $0.11$ and Spearman = $0.14$), indicating the model's ability to find factual error across different contexts is limited. There is no significant difference in factuality between short and long summaries, with error ratios of $16.3\%$ and $15.2\%$, respectively. The system-level ranking for different summary types also showed no significant difference.

In conclusion, our research shows that while LLM-based evaluators perform well in designed factuality datasets, they often generate false positives and struggle with complex, dialogue-based summaries, making them less reliable for automatic factuality evaluation in meeting summarization tasks. Analysis of the IZMS dataset indicates that factuality errors are rarely observed in human annotation feedback and mostly stem from ASR errors in transcripts. in human annotations or out-of-context information from GPT models. Yet GPT models often report out-of-context error, which are false alarms when compared to transcripts. These errors highlight that factuality errors are a rare case in real-world datasets, contrary to the frequency in designed factuality dataset. Hence, when designing our evaluation framework, we prioritize on completeness and conciseness.




\section{CREAM Framework}
\label{sec:creamframe}

To address these issues with existing methods in evaluating long-conetxt summarizations, we propose the CREAM framework, which stands for \underline{C}omparison-Based \underline{R}eference-Free \underline{E}lo-Ranked \underline{A}utomatic Evaluation for \underline{M}eeting Summarization. This framework leverages Chain-of-Thought (CoT) reasoning and comparison-based methods to enhance the evaluation process.

The traditional metrics of completeness and conciseness struggle in this context because they rely heavily on accurate key fact extraction, which is particularly difficult with lengthy, multi-speaker dialogues. The CREAM framework mitigates these challenges by evaluating summarization models through a comparison-based approach that does not require reference to the original transcripts.

As illustrated in Figure \ref{tab:CREAM}, the CREAM framework evaluates two summaries in a reference-free manner, without requiring either gold summary or the document being summarized. CREAM uses a structured, two-step process facilitated by a single CoT-based prompt; details are in Section\ref{sec:prompt}. First, the model extracts concise key facts from a source paragraph, which is created by concatenating both summaries. These key facts must be brief, clear, and non-redundant. In the second step, the model compares these extracted key facts with each summary, assessing whether each fact is accurately reflected and reasoning by identifying the specific sentences in the summary that support it. The results are then parsed for calculation of the summaries' completeness and conciseness, which results in multiple comparison-based scores between models. Detailed results are in Section \ref{sec:comp-score}.

Then, to determine the best models from the comparison-based scores, we introduce an Elo ranking system using comparison-based metrics, as detailed in Section \ref{sec:cream-metrics}. This approach allows for a systematic and robust comparison of model performance by assigning ratings based on pairwise evaluations. The final results are in Section \ref{sec:elo-results}.

\subsection{Implementation}
\subsubsection{Comparison-based Ranking Metrics}
\label{sec:cream-metrics}
We use an reference-free, comparison-based system on each metric in the CREAM framework. The advantage of using comparison-based scores lies in their ability to better differentiate and rank model performance, especially when dealing with the subtle nuances of summarizing lengthy and complex dialogues. This method also enables more effective comparisons when evaluating the impact of different prompts on summary generation. We use ELO system to rank pair-wise comparison results from GPT-4o for both completeness and conciseness.

ELO rating formula can be written as:
$$
R_{new} = R_{old} + K \times (S - E)
$$

where \( R_{new} \) is the new Elo rating after the match; \( R_{old} \) is the old Elo rating before the match. \( K \) is the K-factor, which determines the maximum possible adjustment per match. \( S \) is the actual score (1 for a win, 0.5 for a draw, 0 for a loss). \( E \) is the expected score, calculated as:
$$E = \frac{1}{1 + 10^{(R_{opponent} - R_{old}) / 400}}$$

where \( R_{opponent} \) is the Elo rating of the opponent.

By using Elo rankings instead of raw scores, CREAM focuses on which model performs better in direct comparisons, allowing a reference-free evaluation. This method is robust against missing content, as it evaluates relative performance without relying on comparisons to gold annotations. 

\subsubsection{Prompting Instruction}
\label{sec:prompt}

\input{table/num_key_facts}

We employ CoT zero-shot prompting methods to calculate both conciseness and completeness. This process involves two key steps: first, extracting a set of key facts from the concatenated summaries, and second, comparing these key facts to the summary that is being evaluated.

The prompt instructs the model to perform the following tasks:

1. Key Fact Extraction: The model reads the provided paragraph and breaks it down into a list of key facts. These key facts should be concise, non-repetitive, and should focus on including at most 2-3 entities per fact.

2. Key Fact Alignment: After extracting the key facts, the model compares each fact with the sentences in the summary. For each key fact, the model determines whether it is supported by the summary and, if so, identifies the relevant summary sentences by their line numbers.

The design of the prompt is enhanced by several key techniques, as outlined below:

\textbf{Chain-of-Thought Reasoning }
The prompt utilizes a chain-of-thought approach, incorporating a reasoning step to guide the model through the task. This structured reasoning process helps ensure that the model evaluates the key facts systematically and accurately.

\textbf{One-Shot Learning of Key Facts } 
We experiment with using out-of-domain versus in-domain examples for key fact extraction. Interestingly, out-of-domain examples generally perform better. However, as more in-domain examples were crafted, failure cases increase, with the model occasionally mistaking the example key facts for actual key facts -- a phenomenon observed, though less frequently, with out-of-domain examples as well.

\textbf{Maximum Number of Key Facts }
We test the model’s performance using different limits on the number of key facts, specifically comparing 16 versus 30 key facts. As shown in Table \ref{tab:key-fact-extracted}, GPT-4o is highly sensitive to the number of key facts, while GPT-4 and GPT-3.5 are less sensitive to such instructions. We observe that GPT-3.5 is less stable in consistently extracting the same number of key facts, even for the same summary.

Based on experiments with the IZMS datasets, we decided to use 16 key facts (following precedent literature) and 30 key facts (the maximum effective number for GPT-4o), with GPT-4o as evaluator model, in all subsequent experiments.



\input{table/cream-raw-both}
\input{table/cream-zoom}

\subsection{Experiment Results}
We use the same setups as in Section \ref{sec:setup}, focusing on evaluting completeness and conciseness.
\subsubsection{Pair-wise Raw Scores}
\label{sec:comp-score}

These raw scores of model comparisons show significant differences that are often difficult to discern when using absolute values to compare LLM-generated summaries of long-context dialogues. Table \ref{tab:raw-cream-both} and Table \ref{tab:raw-cream-zoom}  present the raw results of the model comparison scores for completeness and conciseness, on two meeting summarization dataset QMSum and IZMS respectively. For instance, row GPT-4o and column GPT-3.5 means that when GPT-4o and GPT-3.5 are compared pairwise, GPT-3.5 is getting a completeness score of 92.1\%, and GPT-4o is getting a completeness score of 84.2\% (row GPT-3.5 and column GPT-4o).


\paragraph{Interpretable Key Facts}
The intermediate results -- the extracted key facts lists -- provide \textbf{interpretable} insights on the qualitative differences between two summaries. These lists reveal which key facts are consistently captured or missed, offering a deeper understanding of how each model processes and condenses the original content. For instance, in two summaries of a meeting with casual chit-chat followed by important agenda items, Summary A includes details of the chit-chat but omits key decisions made later, while Summary B skips the chit-chat and focuses on the critical decisions. The comparison-based score highlights that although Summary A is more complete in capturing all dialogue, Summary B is more concise and centered on the essential content, making it more valuable for users who need to quickly grasp the meeting outcomes.

\subsubsection{Elo-ranked Results}
\label{sec:elo-results}
As shown in \ref{tab:elo-results} Table CREAM outperforms prior baselines in ranking correlation, especially on meeting summarization data, improving ranking correlation from $0.5$ to $1.0$ (Pearson's r) in both completeness and conciseness. We demonstrate the effectiveness of CREAM in evaluating across different models and identifying optimal prompts for meeting summarization tasks. 
 
We applied the Elo rating system described in Section \ref{sec:creamframe} to our raw scores from system comparisons above. The results align closely with the human evaluation results based on gold summaries, as shown in Table \ref{tab:elo-results}. We see a perfect correlation in ranking of model preference (rank correlation of $1.0$, Pearson's r), even when the gold human-preference scores are close such as the 2nd and 3rd place for completeness score. This consistency suggests that the Elo rating system is an effective method for ranking models in meeting summarization tasks, offering a reliable way to compare their relative performance.

\input{table/elo-results}

\section{Discussion}

\paragraph{The Trade-off Between Metrics}
In evaluating meeting summarization models, a significant trade-off exists between completeness and conciseness. Models that produce more complete summaries often do so at the expense of conciseness, resulting in outputs that are detailed but potentially overwhelming or redundant. Conversely, striving for conciseness can lead to the omission of important details, making the summary less informative. This trade-off is inherently subjective, depending on the specific needs or preferences of the user.

Determining the optimal balance between completeness and conciseness is challenging and varies by application. In practice, this balance is often tailored to individual user preferences. Our approach to key fact extraction can be leveraged to address this trade-off more effectively. By incorporating additional selection steps, users can customize their summaries to focus on the most relevant information, ensuring that even more concise summaries retain all critical facts. Furthermore, this method can help improve model outputs by refining more complete but verbose summaries into concise versions that still capture all essential details.

\paragraph{Benefits in Application}


Our proposed evaluation framework offers several practical benefits, including cost efficiency by using fewer tokens and not requiring the parsing of original transcripts, making the process more affordable. It also enables faster model comparisons, speeding up iterations and refinements. Additionally, as a reference-free evaluation, it eliminates the need for annotations or transcripts, which is especially useful with confidential data, ensuring data privacy and broad applicability. Furthermore, the framework’s adaptability makes it suitable for integration with reinforcement learning (RL) setups.

In future, the prompt used in our framework can be customized to focus on specific sets of key facts, offering significant flexibility in production settings. This adaptability allows users to tailor the evaluation criteria to meet specific needs, such as emphasizing aspects of a summary that are most relevant to the intended audience or application. For example, users could apply filters to key facts, prioritizing main takeaways while ignoring less critical sections like greetings. This level of customization enhances the framework’s utility in diverse real-world scenarios, making it a powerful tool for automated summarization evaluation.

Additionally, we could develop an Elo-based arena based on this work specifically for evaluating summarization models and prompts, allowing for more dynamic and competitive assessments.

\paragraph{Rare Case of Factuality Error}
In our evaluation on real-world meeting data IZMS, we encountered challenges related to factuality errors, but these errors manifested differently between human-annotated gold summaries and GPT-based model outputs. As discussed in Section \ref{sec:Factualityfindings}, these differences suggests a potential issue with divergence in designed datasets versus in application. 

Given these observations, we chose not to specifically target factuality because factuality errors appear to be rare in real-world datasets, unlike their prevalence in designed factuality datasets. As a result, models that claim superior factuality detection performance on such designed datasets may not perform as well in real-world applications, potentially leading to high rates of false positives. To better assess the true degree of factuality errors, we recommend exploring future directions that involve more context-aware evaluation frameworks. 


\section{Conclusions}


In conclusion, we have made several significant contributions to the field of automatically evaluating complex, long-context dialogue summarization. We are the first to examine the effectiveness of LLM-based evaluators for meeting summarization, revealing that general-purpose LLM-based evaluators do not adapt well to the unique challenges posed by long-context dialogue summarization. To address these challenges, we introduced a novel automatic evaluation framework, CREAM, which overcomes the limitations of existing methods by providing a more accurate and adaptable evaluation process using Elo-based comparison algorithms (improving rank correlation of $0.5$, Pearson's r to $1.0$). This framework enables faster and cheaper analysis of models, which can significantly speed up the progress of the field by allowing new methods to be benchmarked more efficiently.

The broader impact of this work extends beyond immediate improvements in evaluation techniques. As models become increasingly sophisticated, human evaluators may struggle to timely assess superhuman AI outputs, highlighting the need for robust, automated evaluation frameworks like CREAM. In the long run, this line of research aims to democratize access to advanced model evaluation, reducing barriers to entry and fostering innovation across the field. Moreover, by providing a scalable and reliable evaluation method, our framework could one day be integrated into reinforcement learning pipelines, offering automatic rewards based on summarization quality -- a critical component given the rising popularity of RL approaches. Ultimately, the adoption of CREAM in industrial applications could help shape the future of AI-driven communication tools, making them more efficient, reliable, and aligned with human needs.
\clearpage

\section*{Acknowledgments}
The authors would like to thank Yanda Chen for valuable feedback and insight to this paper. 

\section*{Limitations}
This work adheres to the ACL Code of Ethics, with all user data appropriately anonymized to protect privacy. However, there are several limitations to consider. The internal data used in our experiments cannot be published due to confidentiality agreements and privacy considerations, which limited our choices in showing some real examples or case studies. Additionally, our evaluations were conducted exclusively using GPT-4o within the CREAM framework. While GPT-4o offers advanced natural language understanding and generation capabilities, it may not be universally applicable as an evaluator, and its assessments could be influenced by biases inherent in the model’s training data. Future work should explore using diverse evaluators, including other models and human assessments, to validate the robustness and generalizability of our findings. These limitations underscore the need for further research and more comprehensive testing to enhance the reliability and applicability of the conclusions drawn from this work.

\clearpage
\bibliography{anthology,custom}
\clearpage
\appendix

\section{Appendix}
\label{sec:appendix}
\subsection{Metrics}
\label{sec:appendix_metrics}

In this study, we follow the definitions and calculation methods outlined by \citet{song2024finesure}, ensuring consistency with recent work in the field. The key metrics used are:

\paragraph{Completeness}
We measure completeness by calculating the proportion of key facts \footnote{A key fact refers to a concise sentence conveying a single key information, comprising at most 2-3 entities, also referred to as a semantic content unit \cite{bhandari-etal-2020-evaluating, song2024finesure}.} from the original text that are accurately captured in the summary. The formula:
$$
\text{Completeness}(K, S, E) = \frac{|(k, s) \in E|}{|K|}
$$
where $K$ represents the set of key facts, $S$ represents the summary, and $E$ is the set of aligned key facts and summary sentences.

\paragraph{Conciseness}

Conciseness is a crucial metric in summarization as it measures the ability of the summary to convey the essential information from the original text without including unnecessary details. A concise summary ensures that the most important points are communicated effectively and efficiently, making it easier for readers to grasp the main ideas without sifting through extraneous content. This is particularly important in meeting summarizations, where lengthy transcripts need to be distilled into clear and focused summaries. We calculate conciseness in a manner similar to completeness, using key facts. The formula is:
$$\text{Conciseness}(K, S, E) = \frac{|s \mid (k, s) \in E|}{|S|}$$
where $K$ represents the set of key facts,  $S$ represents the summary, and $E$ is the set of aligned key facts and summary sentences.

\paragraph{Factuality}
Another critical aspect of evaluating meeting summaries is factuality. To evaluate factuality, we use a CoT approach and calculated a factuality score, defined as:
$$\text{Faithfulness}(D, S) = \frac{|S_{faithful} \in S|}{|S|}$$
where $D$ is given documents, $S$ is the set of sentences in the summary, and $S_{faithful}$ is the sentences in $S$ marked with no factuality errors.

We follow the prompts proposed by \citet{song2024finesure} when evaluating these metrics with baseline methods.

\subsection{Prompt for CREAM}
Our prompt design is attached below:

You will be provided with a paragraph and a summary. Your task is to decompose the paragraph into a set of "key facts". A "key fact" is a single fact written as briefly and clearly as possible, encompassing at most 2-3 entities. There should not be any repeated "key fact". 
Here are nine examples of key facts to illustrate the desired level of granularity (please be careful that these are examples, and do NOT use them as actual key facts):
* Kevin Carr set off on his journey from Haytor.
* Kevin Carr set off on his journey from Dartmoor.
* Kevin Carr set off on his journey in July 2013.
* Kevin Carr is less than 24 hours away from completing his trip.
* Kevin Carr ran around the world unsupported.
* Kevin Carr ran with his tent.
* Kevin Carr is set to break the previous record.
* Kevin Carr is set to break the record by 24 hours.
* The previous record was held by an Australian.
Instruction:
First, read the paragraph carefully. 
Second, decompose the paragraph into (at most 30) ket facts. 

Paragraph:
[INPUT REFERENCE SUMMARIES]

You now have a summary and a set of key facts for the same transcript. Your task is to assess if each key fact is inferred from the summary.

Instruction:
First, compare each key fact with the summary.
Second, check if the key fact is inferred from the summary and then response "Yes" or "No" for each key fact. If "Yes", specify the line number(s) of the summary sentence(s) relevant to each key fact. 

Provide your answer in JSON format. The answer should be a list of dictionaries whose keys are "key fact", "response", and "line number":
[{"key fact": "first key fact", "response": "Yes", "line number": [1]}, {"key fact": "second key fact", "response": "No", "line number": []}, {"key fact": "third key fact", "response": "Yes", "line number": [1, 2, 3]}]

Summary:
[INPUT CANDIDATE SUMMARY]



 

\end{document}

%% file: table/completeness-conciseness-baseline.tex
\begin{table*}[ht]
\centering
\resizebox{0.8\textwidth}{!}{%
\newcommand{\dummy}{\rule[-1em]{0pt}{3em}}
\begin{tabular}{cccccccc}
\hline
\multicolumn{1}{l}{} &
  \begin{tabular}[c]{@{}c@{}}Max\# \\ keyfacts\end{tabular} &
  \begin{tabular}[c]{@{}c@{}}Summarizer\\ Model \end{tabular} &
  \begin{tabular}[c]{@{}c@{}}Human Summary\\ (GPT4o)\end{tabular} &
  \begin{tabular}[c]{@{}c@{}}Machine summary\\ (GPT4o)\end{tabular} &
  \begin{tabular}[c]{@{}c@{}}Transcript\\ (GPT4o)\end{tabular} &
  \begin{tabular}[c]{@{}c@{}}Transcript\\ (GPT4)\end{tabular} &
  \begin{tabular}[c]{@{}c@{}}Transcript\\ (GPT3.5)\end{tabular}\\ \cline{2-8} 
\multicolumn{1}{c|}{\multirow{6}{*}[1em]{\rotatebox[origin=c]{90}{\textbf{COMPLETENESS}}}} &
  \multirow{3}{*}{16} &
  GPT3.5 &
  56.2\% &
  100.0\% &
  89.7\% &
  99.3\% &
  \textbf{99.9\%} \\ \cline{3-8} 
\multicolumn{1}{c|}{} &
   &
  GPT4 &
  57.9\% &
  100.0\% &
  88.5\% &
  99.0\% &
  99.8\% \\ \cline{3-8} 
\multicolumn{1}{c|}{} &
   &
  GPT4o &
  \textbf{77.9\%} &
  100.0\% &
  \textbf{98.9\%} &
  \textbf{99.9\%} &
  99.8\% \\ \cline{2-8} 
\multicolumn{1}{c|}{} &
  \multirow{3}{*}{30} &
  GPT3.5 &
  54.8\% &
  100.0\% &
  82.5\% &
  99.4\% &
  \textbf{99.9\%} \\ \cline{3-8} 
\multicolumn{1}{c|}{} &
   &
  GPT4 &
  55.6\% &
  100.0\% &
  81.2\% &
  99.4\% &
  98.9\% \\ \cline{3-8} 
\multicolumn{1}{c|}{} &
   &
  GPT4o &
  \textbf{76.9\%} &
  100.0\% &
  \textbf{98.1\%} &
  \textbf{99.9\%} &
  99.7\% \\ \hline
\multicolumn{1}{c|}{\multirow{6}{*}{\rotatebox[origin=c]{90}{\textbf{CONCISENESS}}}} &
  \multirow{3}{*}{16} &
  GPT3.5 &
  58.2\% &
  93.3\% &
  91.9\% &
  \textbf{99.0\%} &
  96.7\% \\ \cline{3-8} 
\multicolumn{1}{c|}{} &
   &
  GPT4 &
  \textbf{69.3\%} &
  \textbf{95.0\%} &
  \textbf{96.0\%} &
  \textbf{99.0\%} &
  \textbf{97.8\%} \\ \cline{3-8} 
\multicolumn{1}{c|}{} &
   &
  GPT4o &
  48.8\% &
  62.1\% &
  70.5\% &
  91.2\% &
  78.9\% \\ \cline{2-8} 
\multicolumn{1}{c|}{} &
  \multirow{3}{*}{30} &
  GPT3.5 &
  56.3\% &
  99.4\% &
  97.6\% &
  \textbf{99.4\%} &
  97.1\% \\ \cline{3-8} 
\multicolumn{1}{c|}{} &
   &
  GPT4 &
  \textbf{68.1\%} &
  \textbf{99.8\%} &
  \textbf{98.0\%} &
  99.2\% &
  \textbf{97.5\%} \\ \cline{3-8} 
\multicolumn{1}{c|}{} &
   &
  GPT4o &
  49.8\% &
  92.5\% &
  92.5\% &
  96.3\% &
  79.5\% \\ \hline
\end{tabular}%
}
\caption{Evaluation of current evaluators on completeness and conciseness for meeting summarization. All models show weak correlation with human scores and high self-bias, often assigning near-perfect scores to their own summaries. Summaries generated by various models are evaluated using different evaluator models (in brackets) with varying reference setups. The “Max \# Key Facts” column sets the max number of key facts extracted from reference text. “Human Summary” column is the gold standard baseline using key facts from human summaries.}
\label{tab:completeness-conciseness-baseline-table}
\end{table*}



%% file: table/num_key_facts.tex
\begin{table}[t]
\centering
\resizebox{\columnwidth}{!}{%
\begin{tabular}{lllllllll}
\textbf{\begin{tabular}[c]{@{}l@{}}Max Number\\  of Key Facts\end{tabular}} &
  \textbf{16} &
  \textbf{20} &
  \textbf{30} &
  \textbf{40} &
  \textbf{50} &
  \textbf{60} &
  \textbf{70} &
  \textbf{100} \\ \hline
GPT4o   & 15.8 & 19.7 & 27.0 & 26.6 & 19.6 & 21.3 & 23.4 & 22.4 \\
GPT4   & 12.8 & 12.9 & 13.0 & 12.9 & 12.3 & 12.8 & 13.3 & 12.0 \\
GPT3.5 & 12.4 & 12.3 & 13.2 & 12.5 & 12.7 & 12.9 & 12.6 & 13.3
\end{tabular}%
}
\caption{Average number of key facts extracted from transcripts by different models from one meeting. }
\label{tab:key-fact-extracted}
\end{table}

%% file: table/cream-raw-both.tex
\begin{table}[t]
\centering
\resizebox{\columnwidth}{!}{%
\begin{tabular}{ll|lll|lll}
\multicolumn{1}{c}{\textbf{}} &  & \multicolumn{3}{l|}{\textbf{Completeness}} & \multicolumn{3}{c}{\textbf{Conciseness}} \\ \hline
kf                  & VS     & GPT3.5          & GPT4            & GPT4o  & GPT4            & GPT4            & GPT4o           \\ \hline
\multirow{3}{*}{16} & GPT3.5 & -               & \textbf{78.8\%} & 84.2\% & -               & \textbf{90.0\%} & \textbf{87.5\%} \\
                    & GPT4   & 76.6\%          & \textbf{-}      & 86.6\% & \textbf{93.6\%} & \textbf{-}      & \textbf{92.1\%} \\
                    & GPT4o  & \textbf{92.1\%} & \textbf{94.0\%} & -      & \textbf{67.3\%} & \textbf{68.4\%} & \textbf{-}      \\ \hline
\multirow{3}{*}{30} & GPT3.5 & -               & \textbf{80.7\%} & 82.2\% & -               & 99.1\%          & \textbf{96.4\%} \\
                    & GPT4   & 80.0\%          & -               & 86.3\% & \textbf{99.4\%} & \textbf{-}      & \textbf{95.9\%} \\
                    & GPT4o  & \textbf{94.9\%} & \textbf{95.5\%} & -      & 89.7\%          & 88.6\%          & -              
\end{tabular}%
}
\caption{Raw pair-wise score for completeness and conciseness on QMSum, kf denotes max number of key facts set.}
\label{tab:raw-cream-both}
\end{table}


%% file: table/cream-zoom.tex
\begin{table}[t]
\centering
\resizebox{0.7\columnwidth}{!}{%
\begin{tabular}{lllllll}
\multicolumn{1}{c}{\textbf{}} & \multicolumn{3}{c}{\textbf{Completeness}} &  & \multicolumn{2}{c}{\textbf{Conciseness}} \\ \cline{1-4} \cline{6-7} 
kf                & VS    & short & long       &  & short & long \\ \cline{1-4} \cline{6-7} 
\multirow{2}{*}{16} & short & -     & 95.3       &  & -     & 75.6 \\
                    & long  & 98.4  & \textbf{-} &  & 53.5  & -    \\ \cline{1-4} \cline{6-7} 
\multirow{2}{*}{30} & short & -     & 97.7       &  & -     & 92.7 \\
                    & long  & 97.3  & \textbf{-} &  & 80.7  & -   
\end{tabular}%
}
\caption{Raw pair-wise score for completeness and conciseness on IZMS.}
\label{tab:raw-cream-zoom}
\end{table}

%% file: table/elo-results.tex



\begin{table}[t]
\centering
\resizebox{\columnwidth}{!}{%
\begin{tabular}{llll}
Summary Model        & GPT 4o                     & GPT 4                      & GPT 3.5           \\ \hline
Completeness (Gold)  & \textbf{1st}$_{(77.9\%)}$ & 2nd$_{(57.9\%)}$          & 3rd$_{(54.8\%)}$ \\
Completeness (CREAM) & \textbf{1st}               & 2nd                        & 3rd               \\ \hline
Conciseness (Gold)   & 3rd$_{(48.8\%)}$          & \textbf{1st}$_{(69.3\%)}$ & 2nd$_{(58.2\%)}$ \\
Conciseness (CREAM)  & 3rd                        & \textbf{1st}               & 2nd              
\end{tabular}%
}

\caption{Completeness and conciseness ranking from gold human ranking (human scores in brackets) with reference to gold summary, and ranking from CREAM framework in reference-free setting (16 key fact).}
\label{tab:elo-results}
\end{table}